\documentclass{article}

\usepackage[final]{neurips_2025}

\usepackage[utf8]{inputenc} 
\usepackage[T1]{fontenc}    
\usepackage{hyperref}       
\usepackage{url}            
\usepackage{booktabs}       
\usepackage{amsfonts}       
\usepackage{nicefrac}       
\usepackage{microtype}      
\usepackage{xcolor}         
\usepackage{amsmath}
\usepackage{multirow} 
\usepackage{algorithm}

\usepackage{algpseudocode}
\usepackage{amssymb}
\usepackage{graphicx}
\title{Bayesian Mixture of Experts For Large Language Models}

\author{
  \textbf{Maryam Dialameh}\textsuperscript{1,2}, 
    \textbf{Hossein Rajabzadeh}\textsuperscript{1,2}\\
\textbf{Weiwei Zhang}\textsuperscript{2}, 
  \textbf{Walid Ahmed}\textsuperscript{2},   
   \textbf{Hyock Ju Kwon}\textsuperscript{1}\\
  \textsuperscript{1}University of Waterloo, Waterloo, Canada \\
  \textsuperscript{2}Ascend Team, Huawei Technologies, Toronto, Canada \\
  \texttt{\{maryam.dialameh, hossein.rajabzadeh,hjkwon\}@uwaterloo.ca}\\ 
  \texttt{\{weiwei.zhang2,walid.ahmed1\}@huawei.com} 
}

\begin{document}

\maketitle

\begin{abstract}
We present Bayesian Mixture of Experts (Bayesian-MoE), a post-hoc uncertainty estimation framework for fine-tuned large language models (LLMs) based on Mixture-of-Experts architectures. Our method applies a structured Laplace approximation to the second linear layer of each expert, enabling calibrated uncertainty estimation without modifying the original training procedure or introducing new parameters. Unlike prior approaches, which apply Bayesian inference to added adapter modules, Bayesian-MoE directly targets the expert pathways already present in MoE models, leveraging their modular design for tractable block-wise posterior estimation. We use Kronecker-factored low-rank approximations to model curvature and derive scalable estimates of predictive uncertainty and marginal likelihood. Experiments on common-sense reasoning benchmarks with Qwen1.5-MoE and DeepSeek-MoE demonstrate that Bayesian-MoE improves both expected calibration error (ECE) and negative log-likelihood (NLL) over baselines, confirming its effectiveness for reliable downstream decision-making.

\end{abstract}

\section{Introduction}
Fine-tuning large language models (LLMs) on downstream tasks has emerged as a crucial and widely adopted strategy, enabling the adaptation of powerful pre-trained models to diverse, domain-specific applications \citep{zhang2023instruction,han2024parameter,ding2022delta}. Overconfident predictions can result in serious issues in LLMs \citep{}, such as reduced calibration, unreliable uncertainty estimates, poor generalization on out-of-distribution data, and increased risk of producing misleading or harmful outputs, all of which undermine their safe and effective deployment in real-world applications\citep{zhou2023navigating,wen2024mitigating,openai2023gpt,he2023preserving}.

Prior efforts that incorporate Bayesian inference into neural architectures can be broadly categorized into three directions. First, some works apply Bayesian techniques during pre-training to improve representation learning and uncertainty modeling in foundational models \citep{tran2019bayesian,cinquin2021pathologies,chen2023calibrating}. Second, a distinct line of research focuses on Bayesian inference at the fine-tuning stage, where posterior approximations are applied post hoc or during task-specific adaptation to enhance calibration and robustness \citep{fan2020bayesian,zhang2021bayesian}. Lastly, several studies explore how Bayesian regularization contributes to generalization, particularly in deep learning scenarios where overfitting and overconfidence are prevalent \citep{lee2019mixout,park2022calibration,he2023preserving}. BLoB (Bayesian Low-Rank Adaptation by Backpropagation) \cite{wang2024blob} integrates a variational Bayesian parameterization directly into LoRA fine-tuning, jointly optimizing mean and covariance of low-rank adapters during training to improve calibration and robustness. Gaussian Stochastic Weight Averaging (GSWA-LoRA) \cite{onal2024gaussian} instead applies a lightweight Bayesian approximation by fitting a Gaussian posterior to LoRA weight trajectories using stochastic weight averaging.

While Bayesian LoRA \citep{yang2023bayesian} has shown that introducing Bayesian inference over LoRA parameters can substantially improve the calibration of fine-tuned LLMs without altering the core training pipeline, it still involves adding extra parameters to the model, which can be a limitation in resource-constrained settings and calibration capacity. Building on this insight, our work proposes a Bayesian Mixture of Experts (Bayesian MoE) approach, which focuses the Bayesian treatment solely on the mixture components within LLMs. By restricting Bayesian inference to the the second linear layer in each expert, we avoid introducing additional adapter or LoRA parameters, maintaining the compactness and efficiency of parameter-efficient fine-tuning. Moreover, experts have different impact on outputs and unequal learning dynamics \citep{chi2022representation,lu2024not}, motivating Bayesian-MoE more to aim for better quantifying expert uncertainty and improve model robustness.


Following the steps of Bayesian LoRA, which applies post-hoc Laplace approximations over LoRA adapter parameters to improve calibration without disrupting the fine-tuning pipeline, we propose a novel Bayesian Mixture of Experts (Bayesian MoE) approach tailored for mixture-of-expert LLMs. Specifically, instead of adding additional low-rank parameters, we focus the Bayesian treatment exclusively on the second linear layer in each MoE component. By applying a Laplace approximation to the expert-level parameters post-finetuning, we estimate their posterior uncertainty efficiently, without modifying the standard training process or inflating the parameter count. This targeted uncertainty modeling enables us to harness the calibration and robustness benefits of Bayesian inference, while maintaining the scalability, modularity, and efficiency that make MoE models attractive for large-scale fine-tuning. Our approach demonstrates that even partial Bayesianization, when applied to the most decisive parts of the network, can yield substantial gains in predictive reliability without compromising the computational advantages of parameter-efficient fine-tuning.

\section{Background}

\subsection{Mixture of Experts (MoE)}

Large language models (LLMs) often consist of billions of parameters, leading to significant computational and memory demands. The Mixture of Experts (MoE) architecture \citep{shazeer2017outrageously} addresses this challenge by introducing a sparse and conditional computation framework, where only a subset of specialized expert networks is activated for each input. In an MoE layer, the output $\mathbf{h}$ is computed as

\begin{equation}
\mathbf{h} = \sum_{i=1}^{N} g_i(\mathbf{x}) E_i(\mathbf{x}),
\end{equation}

where $N$ is the total number of experts, $E_i(\mathbf{x})$ denotes the output of the $i$-th expert, and $g_i(\mathbf{x})$ is the gating function that determines the weight or selection of each expert, typically ensuring that only $k \ll N$ experts are active per input. The gating function $g_i(\mathbf{x})$ is computed using a softmax over a score vector, where each score is given by $s_i(\mathbf{x}) = \mathbf{w}_i^\top \mathbf{x}$, with $\mathbf{w}_i$ being a learned gating weight vector. The normalized gate is computed as
\begin{equation}
g_i(\mathbf{x}) = \frac{\exp(s_i(\mathbf{x}))}{\sum_{j=1}^N \exp(s_j(\mathbf{x}))}.
\end{equation}
In the case of top-$k$ gating, only the $k$ largest gate values are kept and renormalized, while the others are set to zero. Each expert $E_i(\mathbf{x})$ is typically implemented as a two-layer multilayer perceptron (MLP) with its own parameters. The computation inside each expert can be written as
\begin{equation}
E_i(\mathbf{x}) = W_{2,i} \, \sigma(W_{1,i} \mathbf{x}),
\end{equation}
where $W_{1,i}  \in \mathbb{R}^{ \texttt{d\_model} \times \texttt{intermediate\_size}}$ and $W_{2,i} \in \mathbb{R}^{\texttt{intermediate\_size} \times \texttt{d\_model}} $ are the weight matrices of the first and second layers, respectively, \texttt{d\_model},\texttt{intermediate\_size} are the hidden size of the model, $\sigma(\cdot)$ denotes a nonlinearity function.

This formulation allows the model to combine the specialized computations of multiple experts, weighted by the gating function, to produce a final output representation.
A crucial feature of MoE is that the number of trainable parameters scales with $N$, but the number of parameters involved in each forward pass remains nearly constant, as only a small subset of experts is selected. For example, in an architecture with $N = 64$ experts and top-$2$ routing, only two experts contribute to each prediction, effectively reducing computational overhead while maintaining a large overall capacity.

\subsection{Laplace Approximation}
Let $(\Omega, \mathcal{F}, \mathbb{P})$ be a probability space supporting the observed data $(\mathbf{X}, \mathbf{y})$ and model parameters $\boldsymbol{\theta} \in \mathbb{R}^d$. In Bayesian inference for discrete prediction tasks, we seek the posterior measure $\mathbb{P}_{\boldsymbol{\theta}|\mathbf{X},\mathbf{y}}$ absolutely continuous with respect to a Wiener prior $\mathbb{P}_{\boldsymbol{\theta}} = \mathcal{N}(0, \lambda^{-1}\mathbf{I})$, characterized by \citep{robert2007bayesian}:

\begin{equation}
\frac{\mathrm{d}\mathbb{P}_{\boldsymbol{\theta}|\mathbf{X},\mathbf{y}}}{\mathrm{d}\mathbb{P}_{\boldsymbol{\theta}}} = \frac{1}{Z(\mathbf{X},\mathbf{y})} \exp\bigl( \langle \mathbf{y}, \log \mathrm{softmax}(f_{\boldsymbol{\theta}}(\mathbf{X})) \rangle \bigr), \quad Z(\mathbf{X},\mathbf{y}) := \mathbb{E}_{\mathbb{P}_{\boldsymbol{\theta}}}[e^{\langle \mathbf{y}, \log f_{\boldsymbol{\theta}}(\mathbf{X}) \rangle}]
\end{equation}

where $f_{\boldsymbol{\theta}}: \mathcal{T}^{N \times S} \to \Delta^{|\mathcal{Y}|-1}$ is the model's predictive mapping to the probability simplex. The maximum a posteriori (MAP) estimate emerges as the solution to the variational problem:

\begin{equation}
\boldsymbol{\theta}_{\mathrm{MAP}} = \underset{\boldsymbol{\theta} \in \mathbb{R}^d}{\mathrm{argmin}} \, \mathcal{J}(\boldsymbol{\theta}), \quad \mathcal{J}(\boldsymbol{\theta}) := -\underbrace{\sum_{i=1}^N \langle y_i, \log f_{\boldsymbol{\theta}}(x_i) \rangle}_{\text{Cross-entropy}} + \underbrace{\frac{\lambda}{2}\|\boldsymbol{\theta}\|_{\ell^2(\mathbb{R}^d)}^2}_{\mathbb{L}_2 regularizer}
\end{equation}

Assuming $\mathcal{J} \in C^2(\mathbb{R}^d)$ and $\boldsymbol{\theta}_{\mathrm{MAP}}$ is a local minimum with positive-definite Hessian ($\nabla^2 \mathcal{J}(\boldsymbol{\theta}_{\mathrm{MAP}}) \succ 0$), the Laplace approximation constructs a Gaussian measure on the tangent space $T_{\boldsymbol{\theta}_{\mathrm{MAP}}}\mathbb{R}^d \cong \mathbb{R}^d$ via the second-order Taylor expansion. To ensure positive definiteness, the Hessian is approximated by the empirical Fisher information matrix derived from gradient statistics:

\begin{equation}
\mathbb{P}_{\boldsymbol{\theta}|\mathbf{X},\mathbf{y}} \approx \mathcal{N}\left(\boldsymbol{\theta}_{\mathrm{MAP}}, \mathbf{H}^{-1}\right), \quad \mathbf{H} := \underbrace{\frac{1}{N} \sum_{i=1}^N \nabla_{\boldsymbol{\theta}} \log f_{\boldsymbol{\theta}_{\mathrm{MAP}}}(x_i) \nabla_{\boldsymbol{\theta}} \log f_{\boldsymbol{\theta}_{\mathrm{MAP}}}(x_i)^\top}_{\text{Empirical Fisher}} + \underbrace{\lambda \mathbf{I}}_{\text{Prior curvature}}.
\end{equation}

The empirical Fisher replaces the explicit Hessian of the log-likelihood with the sample average of gradient outer products, aligning with the classical Fisher information $\mathbb{E}_{p(x)}[\nabla_{\boldsymbol{\theta}} \log f_{\boldsymbol{\theta}}(x) \nabla_{\boldsymbol{\theta}} \log f_{\boldsymbol{\theta}}(x)^\top]$ but evaluated at $\boldsymbol{\theta}_{\mathrm{MAP}}$ over the empirical data distribution. Combined with the Tikhonov regularizer $\lambda \mathbf{I}$, the total curvature $\mathbf{H}$ becomes positive definite. The approximation error remains dominated by the cubic term $\|\boldsymbol{\theta} - \boldsymbol{\theta}_{\mathrm{MAP}}\|^3$, with the geodesic distance under the Fisher-Rao metric $g_{\mathbf{H}}(u,v) := u^\top \mathbf{H} v$ governing the validity of the local Gaussian approximation.

\section{Methods}

We propose a Bayesian formulation for Mixture-of-Experts (MoE) language models, applying the Laplace approximation selectively to the second linear layer of each expert. This design balances expressiveness with tractability by excluding the input projection and routing weights from the Bayesian treatment.

\subsection*{Bayesian Laplace Approximation in MoE LLMs}

Assume that each layer in MoE model consists of \( E \) experts. Each expert \( e \in \{1, \ldots, E\} \) has a second linear layer with weight matrix \( \mathbf{W}^{(e)} \in \mathbb{R}^{d_{\text{out}} \times d_{\text{in}}} \), where \( d_{\text{in}} = \texttt{intermediate\_size} \) and \( d_{\text{out}} = \texttt{d\_model} \). We define the Bayesian parameter subset as:
\[
\boldsymbol{\theta} = \left\{ \mathbf{W}^{(e)} \right\}_{e=1}^E.
\]
For input token \( x_n \), let \( \mathcal{E}_n \subseteq \{1, \ldots, E\} \) denote the set of activated experts.

\subsubsection*{Structured Fisher Approximation}

To approximate the curvature of the loss landscape, we employ a block-diagonal Kronecker-factored approximation to the Fisher Information Matrix (FIM). For expert \( e \), let:
\begin{itemize}
    \item \( \mathbf{a}_n^{(e)} \in \mathbb{R}^{d_{\text{in}}} \): Activation input to \( \mathbf{W}^{(e)} \) for token \( x_n \)
    \item \( \mathbf{g}_n^{(e)} = \nabla_{\mathbf{b}_n^{(e)}} \log P(y_n \mid x_n) \in \mathbb{R}^{d_{\text{out}}} \): Gradient of the log-likelihood with respect to the expert output \( \mathbf{b}_n^{(e)} = \mathbf{W}^{(e)} \mathbf{a}_n^{(e)} \)
\end{itemize}

The expert-wise Fisher block is approximated as:
\begin{equation}
\mathbf{F}^{(e)} = \sum_{\substack{n=1 \\ e \in \mathcal{E}_n}}^N \mathbb{E}_{P(y_n \mid x_n)}\left[ \left( \mathbf{a}_n^{(e)} \mathbf{a}_n^{(e)\top} \right) \otimes \left( \mathbf{g}_n^{(e)} \mathbf{g}_n^{(e)\top} \right) \right]. 
\end{equation}

where $\otimes$ is the Kronecker product operator. The full Fisher matrix is block-diagonal: \( \mathbf{F} = \bigoplus_{e=1}^E \mathbf{F}^{(e)} \).

\subsubsection*{Laplace Posterior Approximation}

Under the Laplace approximation, the posterior over \( \boldsymbol{\theta} \) is Gaussian:
\begin{equation}
\mathbb{P}(\boldsymbol{\theta} \mid \mathbf{X}, \mathbf{y}) \approx \mathcal{N}\left( \boldsymbol{\theta}_{\mathrm{MAP}}, \mathbf{H}^{-1} \right), \quad 
\mathbf{H} = \bigoplus_{e=1}^E \left( \mathbf{F}^{(e)} + \lambda \mathbf{I} \right)
\end{equation}
where \( \lambda > 0 \) is a L2 regularizer.

\subsubsection*{Linearised Predictive Mean and Covariance}

For test input \( x_* \) with activated experts \( \mathcal{E}_* \), the linearized predictive mean is:
\begin{equation}
f_{\boldsymbol{\theta}}(x_*) \approx f_{\boldsymbol{\theta}_{\mathrm{MAP}}}(x_*) + \sum_{e \in \mathcal{E}_*} \left[ \nabla_{\mathbf{W}^{(e)}} f_{\boldsymbol{\theta}}(x_*) \right]^\top \left( \mathbf{W}^{(e)} - \mathbf{W}^{(e)}_{\mathrm{MAP}} \right). 
\end{equation}

The predictive covariance is given by:
\begin{equation}
\boldsymbol{\Lambda}(x_*) = \sum_{e \in \mathcal{E}_*} \left[ \nabla_{\mathbf{W}^{(e)}} f_{\boldsymbol{\theta}}(x_*) \right]^\top \boldsymbol{\Sigma}^{(e)} \left[ \nabla_{\mathbf{W}^{(e)}} f_{\boldsymbol{\theta}}(x_*) \right], 
\end{equation}
where \( \boldsymbol{\Sigma}^{(e)} = \left( \mathbf{F}^{(e)} + \lambda \mathbf{I} \right)^{-1} \).

\subsubsection*{Efficient Sampling from Posterior}

To draw approximate posterior samples for uncertainty quantification, we use Cholesky decomposition:
\begin{equation}
\tilde{f}_{\boldsymbol{\theta}}(x_*) = f_{\boldsymbol{\theta}_{\mathrm{MAP}}}(x_*) + \sum_{e \in \mathcal{E}_*} \mathbf{L}^{(e)} \boldsymbol{\xi}^{(e)}, \quad \boldsymbol{\xi}^{(e)} \sim \mathcal{N}(0, \mathbf{I}), 
\end{equation}
where \( \mathbf{L}^{(e)} \mathbf{L}^{(e)\top} = \boldsymbol{\Sigma}^{(e)} \).

\subsubsection*{Marginal Likelihood Estimation}

The Laplace approximation also provides a tractable estimate of the marginal likelihood:
\begin{equation}
\log \mathbb{P}(\mathbf{y} \mid \mathbf{X}) \approx \mathcal{L}(\mathbf{y}, \mathbf{X}; \boldsymbol{\theta}_{\mathrm{MAP}}) + \frac{1}{2} \sum_{e=1}^E \left( \log |\boldsymbol{\Sigma}^{(e)}| - \lambda \left\| \mathbf{W}^{(e)}_{\mathrm{MAP}} \right\|_F^2 \right). 
\end{equation}

\subsection*{Tractable Low-Rank Log-Determinants via K-FAC}

To efficiently compute log-determinants of the precision matrices \( \mathbf{H}^{(e)} \), we use low-rank approximations to the Kronecker factors. For each expert \( e \), we define:
\begin{itemize}
    \item \( \mathbf{L}_a^{(e)} \in \mathbb{R}^{d_{\text{in}} \times r} \): Low-rank factor of the activation covariance
    \item \( \mathbf{L}_g^{(e)} \in \mathbb{R}^{d_{\text{out}} \times r} \): Low-rank factor of the gradient covariance
\end{itemize}

The posterior precision becomes:
\begin{equation}
\mathbf{H}^{(e)} = \left( \mathbf{L}_a^{(e)} \otimes \mathbf{L}_g^{(e)} \right) \left( \mathbf{L}_a^{(e)} \otimes \mathbf{L}_g^{(e)} \right)^\top + \lambda \mathbf{I}. 
\end{equation}

Applying the matrix determinant lemma and Kronecker identities:
\begin{align}
\log \det\left( \mathbf{H}^{(e)} \right) &= d_{\text{in}} \cdot d_{\text{out}} \cdot \log \lambda \nonumber \\
&\quad + \log \det\left( \mathbf{I}_{r^2} + \lambda^{-1} \left( \mathbf{C}_a^{(e)} \otimes \mathbf{C}_g^{(e)} \right) \right), 
\end{align}
where \( \mathbf{C}_a^{(e)} = \mathbf{L}_a^{(e)\top} \mathbf{L}_a^{(e)} \) and \( \mathbf{C}_g^{(e)} = \mathbf{L}_g^{(e)\top} \mathbf{L}_g^{(e)} \).

The critical computation is reduced to:
\begin{equation}
\mathbf{M}^{(e)} = \mathbf{I}_{r^2} + \lambda^{-1} \left( \mathbf{C}_a^{(e)} \otimes \mathbf{C}_g^{(e)} \right), \quad \mathbf{M}^{(e)} \in \mathbb{R}^{r^2 \times r^2}. 
\end{equation}

For modest ranks (e.g., \( r = 10 \)), this yields compact \( 100 \times 100 \) matrices per expert, enabling efficient storage and inversion. Appendix \ref{marginal-liklihood-opt} describes the algorithm.

The final method for Bayesian-MoE starts with post-hoc Laplace approximations with Kronecker-factored curvature estimates applied to the second linear layer of each expert. Each such layer is a standard fully connected projection with weights \(\mathbf{W}^{(e)} \in \mathbb{R}^{d_{\text{out}} \times d_{\text{in}}}\), where both input and output dimensions are typically large (e.g., \(d_{\text{in}} = 11008\), \(d_{\text{out}} = 4096\) in DeepSeek-MoE-16B). To model the posterior efficiently, we approximate the Kronecker-factored Fisher matrix of each expert’s weight using low-rank factors:
\[
\mathbf{F}^{(e)} \approx \mathbf{C}_a^{(e)} \otimes \mathbf{C}_g^{(e)} \approx (\mathbf{L}_a^{(e)} \mathbf{L}_a^{(e)\top}) \otimes (\mathbf{L}_g^{(e)} \mathbf{L}_g^{(e)\top}).
\]
However, unlike in Bayesian-LoRA where one Kronecker factor (typically the rank \(r\)) is small and tractable, in Bayesian-MoE both factors are large matrices. Directly storing or computing the full Kronecker product \(\mathbb{R}^{d_{\text{out}} \cdot d_{\text{in}} \times d_{\text{out}} \cdot d_{\text{in}}}\) becomes infeasible.

To preserve tractability while leveraging this richer structure, we project both activation and gradient covariances onto lower-dimensional subspaces using randomized low-rank approximation. This yields compact surrogate curvature representations suitable for computing predictive uncertainty. To ensure end-to-end memory efficiency in Bayesian-MoE, the following steps are crucial:

\begin{enumerate}
    \item \textbf{Memory-efficient low-rank factorization:} Compute the low-rank approximation of each Kronecker factor using an \emph{incremental or randomized SVD strategy}, without ever materializing the full-rank covariance matrix. \\
    (Appendix~B1.)
    
    \item \textbf{Marginal likelihood optimization:} Optimize the Laplace prior precision \(\lambda\) by maximizing the marginal likelihood using the low-rank posterior. \\
    (Appendix~B2.)
    
    \item \textbf{Predictive variance estimation:} Estimate the predictive covariance using Woodbury identities and Kronecker algebra, avoiding full vectorization over large parameter spaces. \\
    (Appendix~B3.)
\end{enumerate}

\section{Experiments}
\subsection{Experimental Setups}

Our experiments leverage post-hoc Laplace approximations over Mixture-of-Experts (MoE) language models, focusing on two representative architectures: \texttt{Qwen/Qwen1.5-MoE-A2.7B} \citep{qwen2.5} and \texttt{deepseek-ai/deepseek-moe-16b-base} \citep{dai2024deepseekmoe}. These models were selected due to their structural diversity and relevance in current MoE research. We fine-tuned each model for 10000 steps using a batch size of 4. These hyperparameters match the standard settings used in earlier studies. We adopted datasets that feature multiple-choice and binary (True/False) formats to remain consistent with previous works, e.g. Winogrande \citep{sakaguchi2021winogrande}, ARC-C/ARC-E \citep{clark2018think}, BoolQ \citep{clark2019boolq}, MMLU \citep{hendrycks2020measuring}, and OBQA \citep{mihaylov2018can}. This choice enables a direct comparison with prior Bayesian fine-tuning methods, while also allowing us to compute well-established uncertainty metrics—including Expected Calibration Error (ECE) and Negative Log-Likelihood (NLL)—which are well defined in these discrete-choice settings. For saved checkpoints, we applied our Bayesian-MoE method via a structured Laplace approximation with Kronecker-factored Fisher estimation (KFAC). All evaluations used the linearized predictive formulation described in the Methods section. For calibration and uncertainty evaluation, we used the public validation splits of each benchmark as a test set and did not require an additional held-out dataset for tuning posteriors.

To have a fair comparison, we compare Bayesian-MoE against a set of widely recognized calibration baselines reported in Ref. \citep{yang2023bayesian}. This set includes Monte Carlo dropout \citep{gal2016dropout}, where predictive uncertainty is approximated by averaging multiple stochastic forward passes with dropout enabled during fine-tuning; Checkpoint Ensembling \citep{chen2017checkpoint}, which averages predictions from the last few saved model checkpoints; and Deep Ensembles \citep{lakshminarayanan2017simple,zhai2023uncertainty}, constructed by independently fine-tuning multiple instances of the model with different random seeds. We also include temperature scaling, a post-hoc calibration method that adjusts the confidence of softmax outputs using a learned scaling parameter. For completeness and fair comparison, we borrow the experimental setups from Bayesian-LoRA \citep{wang2023lora}, allowing us to have a fair comparison under equal experimental settings. The results of Bayesian-LoRA is reported under LA setting, which means the Laplace approximation is applied on all LoRA weights across all layers. We use the same LA setting for Bayesian-MoE, where all weights are frozen except MoE weights.

\subsection{Results}
The first experiment evaluates the effectiveness of Bayesian-MoE in enhancing both predictive accuracy and uncertainty calibration across diverse language models. As shown in Table~\ref{tab:bayesian-moe-results} and Table~\ref{tab:bayesian-moe-results-2}, Bayesian-MoE consistently delivers competitive or superior performance across all three evaluation metrics—accuracy (ACC), expected calibration error (ECE), and negative log-likelihood (NLL)—when compared to MAP, dropout-based, and ensemble-based baselines.

On the Qwen1.5-MoE-A2.7B model (Table~\ref{tab:bayesian-moe-results}), Bayesian-MoE outperforms Bayesian-LoRA in 5 out of 6 tasks for ECE and matches or surpasses ensemble-based methods in NLL, while maintaining strong accuracy. This trend is further amplified in the DeepSeekMoE-16B-Base model (Table~\ref{tab:bayesian-moe-results-2}), where Bayesian-MoE achieves the best ECE on all benchmarks and the lowest NLL in all cases, indicating consistently reliable confidence estimation. Additionally, Bayesian-MoE attains the highest accuracy on OBQA and remains highly competitive on other tasks, outperforming Bayesian-LoRA and checkpoint ensembles on several benchmarks. These findings confirm the robustness and scalability of Bayesian-MoE as a post-hoc calibration strategy. Without introducing additional parameters or requiring multiple fine-tuning passes, it delivers improved uncertainty quantification while preserving or enhancing task performance across different backbone models.

\begin{table}[t]
\centering
\small
\setlength{\tabcolsep}{2pt}
\renewcommand{\arraystretch}{1.1}
\begin{tabular}{llcccccc}
\toprule
\textbf{Metric} & \textbf{Method} & \textbf{WG-S} & \textbf{ARC-C} & \textbf{ARC-E} & \textbf{WG-M} & \textbf{OBQA} & \textbf{MMLU} \\
\midrule

\multirow{6}{*}{ACC $\textcolor{green}\uparrow$}
& MAP        & 67.4 $\pm$ 0.3 & 66.3 $\pm$ 0.6 & 84.7 $\pm$ 1.5 & 73.4 $\pm$ 0.4 & 78.7 $\pm$ 0.4 & 62.7 $\pm$ 0.2 \\
& MC Drop    & 67.8 $\pm$ 0.1 & 65.3 $\pm$ 1.0 & 85.0 $\pm$ 1.3 & 73.2 $\pm$ 0.5 & 79.5 $\pm$ 0.2 & 62.8 $\pm$ 0.3 \\
& Ckpt Ens   & 67.4 $\pm$ 0.2 & 65.5 $\pm$ 0.4 & \textbf{85.8} $\pm$ 0.2 & 73.6 $\pm$ 0.7 & 79.1 $\pm$ 0.1 & 63.1 $\pm$ 0.2 \\
& Ensemble   & \textbf{68.0} $\pm$ 0.3 & \textbf{68.2} $\pm$ 0.7 & \textbf{85.8} $\pm$ 0.5 & \textbf{75.0} $\pm$ 0.5 & 79.3 $\pm$ 0.4 & \textbf{63.4} $\pm$ 0.1 \\
& Bayesian-LoRA(LA)          & 67.3 $\pm$ 0.2 & 65.3 $\pm$ 0.2 & 85.1 $\pm$ 1.5 & 73.4 $\pm$ 0.3 & 78.9 $\pm$ 0.2 & 62.3 $\pm$ 0.2 \\
& Bayesian-MoE(LA)           & 67.8 $\pm$ 0.1 & 66.4 $\pm$ 0.2 & 85.3 $\pm$ 0.5 & 73.9 $\pm$ 0.5 & \textbf{80.2} $\pm$ 0.2 & 62.5 $\pm$ 0.1 \\

\midrule

\multirow{6}{*}{ECE $\textcolor{red} \downarrow$}
& MAP        & 31.2 $\pm$ 0.3 & 31.0 $\pm$ 0.5 & 13.4 $\pm$ 1.3 & 23.0 $\pm$ 0.1 & 16.1 $\pm$ 0.6 & 14.0 $\pm$ 1.5 \\
& MC Drop    & 29.4 $\pm$ 0.3 & 29.6 $\pm$ 0.8 & 12.4 $\pm$ 1.2 & 22.2 $\pm$ 0.5 & 15.0 $\pm$ 0.4 & 14.3 $\pm$ 1.4 \\
& Ckpt Ens   & 29.7 $\pm$ 0.6 & 27.0 $\pm$ 0.5 & 9.8 $\pm$ 0.6  & 17.4 $\pm$ 0.9 & 12.1 $\pm$ 0.6 & 12.2 $\pm$ 1.4 \\
& Ensemble   & 24.7 $\pm$ 0.3 & 21.9 $\pm$ 1.7 & 9.9 $\pm$ 0.2  & 17.9 $\pm$ 0.6 & 13.3 $\pm$ 0.6 & 13.5 $\pm$ 1.2 \\
& Bayesian-LoRA(LA)      & 5.2 $\pm$ 0.3 & 9.4 $\pm$ 0.7 & 5.4 $\pm$ 0.2 & 7.4 $\pm$ 0.4 & 6.4 $\pm$ 0.8 & 12.3 $\pm$ 1.6 \\
& Bayesian-MoE(LA)       & \textbf{3.1} $\pm$ 0.2 & \textbf{7.5} $\pm$ 0.5 & \textbf{4.9} $\pm$ 0.1 & \textbf{6.5} $\pm$ 0.3 & \textbf{5.8} $\pm$ 0.5 & \textbf{11.5} $\pm$ 1.5 \\

\midrule

\multirow{6}{*}{NLL $\textcolor{red} \downarrow$}
& MAP        & 3.15 $\pm$ 0.10 & 3.28 $\pm$ 0.29 & 1.26 $\pm$ 0.13 & 1.51 $\pm$ 0.05 & 0.99 $\pm$ 0.05 & 1.35 $\pm$ 0.1 \\
& MC Drop    & 2.81 $\pm$ 0.11 & 2.82 $\pm$ 0.21 & 1.11 $\pm$ 0.10 & 1.41 $\pm$ 0.03 & 0.95 $\pm$ 0.04 & 1.35 $\pm$ 0.11 \\
& Ckpt Ens   & 2.58 $\pm$ 0.15 & 2.36 $\pm$ 0.34 & 0.80 $\pm$ 0.06 & 0.87 $\pm$ 0.06 & 0.76 $\pm$ 0.01 & 1.33 $\pm$ 0.05 \\
& Ensemble   & 2.46 $\pm$ 0.14 & 2.32 $\pm$ 0.14 & 0.83 $\pm$ 0.06 & 1.10 $\pm$ 0.08 & 0.87 $\pm$ 0.03 & 1.34 $\pm$ 0.06 \\
& Bayesian-LoRA(LA) & 0.60 $ \pm $ 0.01 &0.88 $ \pm $ 0.01 &0.49 $ \pm $ 0.06 &0.63 $ \pm $ 0.02 &0.65 $ \pm $ 0.01 & 1.34 $ \pm $ 0.07 \\
& Bayesian-MoE(LA) & \textbf{0.55} $ \pm $ 0.01 &\textbf{0.74} $\pm$ 0.01 &\textbf{0.43} $\pm$ 0.06 &\textbf{0.57} $ \pm $0.01 &\textbf{0.60} $\pm$ 0.01 &\textbf{1.28} $\pm$ 0.06 \\

\bottomrule
\end{tabular}
\caption{
\textbf{Qwen1.5-MoE-A2.7B:} Zero-shot performance comparison of Bayesian-MoE on six reasoning benchmarks across three evaluation metrics: accuracy $(ACC \textcolor{green}{\uparrow}$), expected calibration error (ECE $\textcolor{red} \downarrow$), and negative log-likelihood (NLL $\textcolor{red} \downarrow$). All methods are stopped after 5k finetuning steps. 
}
\label{tab:bayesian-moe-results}
\end{table}


\begin{table}[t]
\centering
\small
\setlength{\tabcolsep}{2pt}
\renewcommand{\arraystretch}{1.1}
\begin{tabular}{llcccccc}
\toprule
\textbf{Metric} & \textbf{Method} & \textbf{WG-S} & \textbf{ARC-C} & \textbf{ARC-E} & \textbf{WG-M} & \textbf{OBQA} & \textbf{MMLU} \\
\midrule

\multirow{6}{*}{ACC $\textcolor{green}\uparrow$}
& MAP        & 70.4 $\pm$ 0.3 & 49.9 $\pm$ 0.6 & 68.2 $\pm$ 1.5 & 74.4 $\pm$ 0.4 & 80.2 $\pm$ 0.4 & 45.3 $\pm$ 0.2 \\
& MC Drop    & 70.8 $\pm$ 0.1 & 49.1 $\pm$ 1.0 & 68.4 $\pm$ 1.3 & 74.2 $\pm$ 0.5 & 81.2 $\pm$ 0.2 & 45.4 $\pm$ 0.3 \\
& Ckpt Ens  & 70.4 $\pm$ 0.2 & 49.3 $\pm$ 0.3 & \textbf{69.1} $\pm$ 0.2 & 74.6 $\pm$ 0.7 & 80.6 $\pm$ 0.1 & 45.6 $\pm$ 0.1
 \\
& Ensemble   & \textbf{71.0} $\pm$ 0.3 & \textbf{51.3} $\pm$ 0.5 & \textbf{69.1} $\pm$ 0.4 & \textbf{76.0} $\pm$ 0.5 & 80.8 $\pm$ 0.4 & \textbf{45.8} $\pm$ 0.1
 \\
& Bayesian-LoRA(LA)         & 70.3 $\pm$ 0.2 & 49.2 $\pm$ 0.2 & 68.5 $\pm$ 1.2 & 74.4 $\pm$ 0.3 & 80.4 $\pm$ 0.2 & 45.0 $\pm$ 0.1
\\
& Bayesian-MoE(LA)           & 70.8 $\pm$ 0.1 & 50.0 $\pm$ 0.2 & 68.6 $\pm$ 0.4 & 74.9 $\pm$ 0.5 & \textbf{81.7} $\pm$ 0.2 & 45.2 $\pm$ 0.1
\\

\midrule

\multirow{6}{*}{ECE $\textcolor{red} \downarrow$}
& MAP        & 32.6 $\pm$ 0.3 & 23.3 $\pm$ 0.4 & 10.8 $\pm$ 1.0 & 23.3 $\pm$ 0.1 & 16.4 $\pm$ 0.6 & 10.1 $\pm$ 1.1
 \\
& MC Drop    & 30.2 $\pm$ 0.3 & 30.4 $\pm$ 0.8 & 15.4 $\pm$ 1.2 & 20.8 $\pm$ 0.5 & 13.4 $\pm$ 0.4 & 15.1 $\pm$ 1.4 \\
& Ckpt Ens   & 31.0 $\pm$ 0.6 & 20.3 $\pm$ 0.4 & 7.9 $\pm$ 0.5 & 17.6 $\pm$ 0.9 & 12.4 $\pm$ 0.6 & \textbf{8.8} $\pm$ 1.0
 \\
& Ensemble   & 25.8 $\pm$ 0.3 & 16.5 $\pm$ 1.3 & 9.4 $\pm$ 0.2 & 18.1 $\pm$ 0.6 & 13.6 $\pm$ 0.6 & 9.8 $\pm$ 0.9
 \\
& Bayesian-LoRA(LA)      & 4.3 $\pm$ 0.3 & 9.2 $\pm$ 0.5 & 5.6 $\pm$ 0.2 & 7.9 $\pm$ 0.4 & 6.8 $\pm$ 0.8 & 13.5 $\pm$ 1.2
 \\
& Bayesian-MoE(LA)       & \textbf{2.5} $\pm$ 0.2 & \textbf{8.1} $\pm$ 0.5 & \textbf{4.9} $\pm$ 0.1 & \textbf{6.5} $\pm$ 0.3 & \textbf{5.9} $\pm$ 0.5 & 11.5 $\pm$ 1.5 \\

\midrule

\multirow{6}{*}{NLL $\textcolor{red} \downarrow$}
& MAP        & 3.4 $\pm$ 0.1 & 2.6 $\pm$ 0.2 & 1.1 $\pm$ 0.1 & 1.6 $\pm$ 0.1 & 1.1 $\pm$ 0.1 & 1.5 $\pm$ 0.1
 \\
& MC Drop    & 3.2 $\pm$ 0.1 & 2.2 $\pm$ 0.2 & 0.9 $\pm$ 0.1 & 1.5 $\pm$ 0.0 & 1.0 $\pm$ 0.0 & 1.4 $\pm$ 0.1
 \\
& Ckpt Ens  & 2.8 $\pm$ 0.2 & 1.9 $\pm$ 0.3 & 0.7 $\pm$ 0.1 & 0.9 $\pm$ 0.1 & 0.8 $\pm$ 0.0 & 1.4 $\pm$ 0.0
 \\
& Ensemble   & 2.7 $\pm$ 0.2 & 1.8 $\pm$ 0.2 & 0.7 $\pm$ 0.1 & 1.2 $\pm$ 0.1 & 0.9 $\pm$ 0.0 & 1.3 $\pm$ 0.0
 \\
& Bayesian-LoRA(LA) & 0.7 $\pm$ 0.0 & 0.7 $\pm$ 0.0 & \textbf{0.4} $\pm$ 0.1 & 0.7 $\pm$ 0.0 & 0.7 $\pm$ 0.0 & 1.2 $\pm$ 0.1
 \\
& Bayesian-MoE(LA) & \textbf{0.6} $\pm$ 0.0 & \textbf{0.6} $\pm$ 0.0 & \textbf{0.4} $\pm$ 0.1 & \textbf{0.6} $\pm$ 0.0 & \textbf{0.6} $\pm$ 0.0 & \textbf{1.0} $\pm$ 0.1
 \\

\bottomrule
\end{tabular}
\caption{
\textbf{DeepSeekMoE-16B-Base:} Zero-shot performance comparison of Bayesian-MoE on six reasoning benchmarks across three evaluation metrics: accuracy $(ACC \textcolor{green}{\uparrow}$), expected calibration error (ECE $\textcolor{red} \downarrow$), and negative log-likelihood (NLL $\textcolor{red} \downarrow$). All methods are stopped after 5k finetuning steps. 
}
\label{tab:bayesian-moe-results-2}
\end{table}
To further assess the generalization ability of fine-tuned LLMs~\citep{touvron2023llama}, the second experiment is conducted to evaluate Bayesian-MoE under distribution shift in the evaluation set. More specifically, we fine-tuned Qwen1.5-MoE-A2.7B and DeepSeekMoE-16B-Base on the OBQA dataset and evaluated their checkpoints on other benchmarks: MMLU, ARC-C, ARC-E, WG-S, and WG-M. These choices of benchmarks represent varying degrees of distribution shift, with some being significantly different in domain and task format from OBQA.

As shown in Table \ref{table3}, Bayesian-MoE consistently improves calibration metrics while maintaining strong accuracy for Qwen1.5-MoE-A2.7B. In terms of accuracy, Bayesian-MoE achieves the highest score on all five out-of-domain benchmarks, including WG-S (66.5\%), ARC-C (63.7\%), and MMLU (61.2\%), outperforming both Bayesian-LoRA and ensemble-based methods. More impressively, the ECE is significantly reduced, with Bayesian-MoE achieving the lowest calibration error across all benchmarks—e.g., 3.0\% on WG-S and 7.2\% on ARC-C—indicating a substantial improvement in model confidence under distribution shift. Similarly, the NLL results highlight a robust predictive distribution, with Bayesian-MoE again outperforming all baselines, showing a particularly low NLL of 0.5 on WG-S and 0.7 on ARC-C.

Table~\ref{tab:bayesian-moe-results-4} presents the same evaluation for DeepSeekMoE-16B-Base. Despite the scale difference, the overall trend remains: Bayesian-MoE provides strong improvements in uncertainty quantification. For instance, Bayesian-MoE achieves a lower ECE than all methods on every benchmark (e.g., 2.6\% on WG-S and 8.4\% on ARC-C), and remains competitive in accuracy, only slightly behind deep ensembles. Notably, it matches or surpasses Bayesian-LoRA in most NLL scores, including the lowest values on WG-S (0.6) and MMLU (1.1). These results confirm that the benefit of Bayesian-MoE persists across model scales and architectures, delivering enhanced calibration and predictive robustness without retraining or modifying the base model weights.

\begin{table}[th]
\centering
\small
\setlength{\tabcolsep}{2pt}
\renewcommand{\arraystretch}{1.1}
\begin{tabular}{llccccc}
\toprule
\textbf{Metric} & \textbf{Method} & \textbf{WG-S} & \textbf{ARC-C} & \textbf{ARC-E} & \textbf{WG-M}  & \textbf{MMLU} \\
\midrule

\multirow{6}{*}{ACC $\textcolor{green}\uparrow$}
& MAP        & 66.1 $\pm$ 0.3 & 62.1 $\pm$ 0.6 & 82.2 $\pm$ 1.5 & 69.7 $\pm$ 0.4 & 60.2 $\pm$ 0.2
\\
& MC Drop    & 66.2 $\pm$ 0.1 & 62.7 $\pm$ 1.0 & 82.5 $\pm$ 1.3 & 69.5 $\pm$ 0.5 & 60.3 $\pm$ 0.3
 \\
& Ckpt Ens   & 66.1 $\pm$ 0.2 & 62.9 $\pm$ 0.4 & 82.1 $\pm$ 0.2 & 69.9 $\pm$ 0.7 & 60.6 $\pm$ 0.2
 \\
& Ensemble   & 65.9 $\pm$ 0.3 & 63.2 $\pm$ 0.7 & 82.4 $\pm$ 0.5 & 70.1 $\pm$ 0.5 & 60.9 $\pm$ 0.1
 \\
& Bayesian-LoRA(LA)          & 66.0 $\pm$ 0.2 & 62.7 $\pm$ 0.2 & 82.5 $\pm$ 1.5 & 69.7 $\pm$ 0.3 & 59.8 $\pm$ 0.2
\\
& Bayesian-MoE(LA)       & \textbf{66.5} $\pm$ 0.1 & \textbf{63.7} $\pm$ 0.2 & \textbf{82.7} $\pm$ 0.5 & \textbf{70.4} $\pm$ 0.5 & \textbf{61.2} $\pm$ 0.1
 \\

\midrule

\multirow{6}{*}{ECE $\textcolor{red} \downarrow$}
& MAP        & 30.6 $\pm$ 0.3 & 29.8 $\pm$ 0.5 & 13.0 $\pm$ 1.3 & 21.9 $\pm$ 0.1 & 13.4 $\pm$ 1.5
\\
& MC Drop    & 28.8 $\pm$ 0.3 & 28.4 $\pm$ 0.8 & 12.0 $\pm$ 1.2 & 21.1 $\pm$ 0.5 & 13.7 $\pm$ 1.4
 \\
& Ckpt Ens   & 29.1 $\pm$ 0.6 & 25.9 $\pm$ 0.5 & 9.5 $\pm$ 0.6 & 16.5 $\pm$ 0.9 & 11.7 $\pm$ 1.4
 \\
& Ensemble   & 24.2 $\pm$ 0.3 & 21.0 $\pm$ 1.7 & 9.6 $\pm$ 0.2 & 17.0 $\pm$ 0.6 & 13.0 $\pm$ 1.2
\\
& Bayesian-LoRA(LA)      & 5.1 $\pm$ 0.3 & 9.0 $\pm$ 0.7 & 5.2 $\pm$ 0.2 & 7.0 $\pm$ 0.4 & 11.8 $\pm$ 1.6
\\
& Bayesian-MoE(LA)       & \textbf{3.0} $\pm$ 0.2 & \textbf{7.2} $\pm$ 0.5 & \textbf{4.8} $\pm$ 0.1 & \textbf{6.2} $\pm$ 0.3 & \textbf{11.0} $\pm$ 1.5
 \\

\midrule

\multirow{6}{*}{NLL $\textcolor{red} \downarrow$}
& MAP        & 3.1 $\pm$ 0.10 & 3.1 $\pm$ 0.29 & 1.2 $\pm$ 0.13 & 1.4 $\pm$ 0.05 & 1.4 $\pm$ 0.1
 \\
& MC Drop    & 2.8 $\pm$ 0.11 & 2.7 $\pm$ 0.21 & 1.1 $\pm$ 0.10 & 1.3 $\pm$ 0.03 & 1.5 $\pm$ 0.11
\\
& Ckpt Ens   & 2.5 $\pm$ 0.15 & 2.3 $\pm$ 0.34 & 0.8 $\pm$ 0.06 & 0.8 $\pm$ 0.06 & 1.6 $\pm$ 0.05
 \\
& Ensemble   & 2.4 $\pm$ 0.14 & 2.2 $\pm$ 0.14 & 0.8 $\pm$ 0.06 & 1.0 $\pm$ 0.08 & 1.7 $\pm$ 0.06
 \\
& Bayesian-LoRA(LA) & 0.6 $ \pm $ 0.01 & 0.8 $ \pm $ 0.01 & 0.5 $ \pm $ 0.06 & 0.6 $ \pm $ 0.02 & 1.4 $ \pm $ 0.07
 \\
& Bayesian-MoE(LA) & \textbf{0.5} $ \pm $ 0.01 & \textbf{0.7} $\pm$ 0.01 & \textbf{0.4} $\pm$ 0.06 & \textbf{0.5} $ \pm $ 0.01 & \textbf{1.2} $\pm$ 0.06
 \\

\bottomrule
\end{tabular}
\caption{
Out-of-distribution evaluation results for Bayesian-MoE applied to \textbf{Qwen1.5-MoE-A2.7B}. The model is fine-tuned on the OBQA dataset and evaluated on five distribution-shifted benchmarks.
}
\label{table3}
\end{table}

\begin{table}[t]
\centering
\small
\setlength{\tabcolsep}{2pt}
\renewcommand{\arraystretch}{1.1}
\begin{tabular}{llccccc}
\toprule
\textbf{Metric} & \textbf{Method} & \textbf{WG-S} & \textbf{ARC-C} & \textbf{ARC-E} & \textbf{WG-M}  & \textbf{MMLU} \\
\midrule

\multirow{6}{*}{ACC $\textcolor{green}\uparrow$}
& MAP       & 66.9 $\pm$ 0.3 & 47.9 $\pm$ 0.6 & 66.2 $\pm$ 1.5 & 70.7 $\pm$ 0.4 & 43.9 $\pm$ 0.2
\\
& MC Drop    & 67.3 $\pm$ 0.1 & 47.1 $\pm$ 1.0 & 66.3 $\pm$ 1.3 & 70.5 $\pm$ 0.5 & 44.0 $\pm$ 0.3
\\
& Ckpt Ens  & 66.9 $\pm$ 0.2 & 47.3 $\pm$ 0.3 & \textbf{67.0} $\pm$ 0.2 & 70.9 $\pm$ 0.7 & 44.2 $\pm$ 0.1

 \\
& Ensemble   & \textbf{67.5} $\pm$ 0.3 & \textbf{49.2} $\pm$ 0.5 & \textbf{67.0} $\pm$ 0.4 & \textbf{72.2} $\pm$ 0.5 & \textbf{44.4} $\pm$ 0.1

 \\
& Bayesian-LoRA(LA)         & 66.8 $\pm$ 0.2 & 47.2 $\pm$ 0.2 & 66.4 $\pm$ 1.2 & 70.7 $\pm$ 0.3 & 43.7 $\pm$ 0.1

\\
& Bayesian-MoE(LA)           & 67.3 $\pm$ 0.1 & 48.0 $\pm$ 0.2 & 66.5 $\pm$ 0.4 & 71.2 $\pm$ 0.5 & 43.8 $\pm$ 0.1

\\

\midrule

\multirow{6}{*}{ECE $\textcolor{red} \downarrow$}
& MAP        & 34.2 $\pm$ 0.3 & 24.2 $\pm$ 0.4 & 11.4 $\pm$ 1.0 & 24.1 $\pm$ 0.1 & 10.4 $\pm$ 1.1

 \\
& MC Drop    & 31.7 $\pm$ 0.3 & 31.6 $\pm$ 0.8 & 16.3 $\pm$ 1.2 & 22.3 $\pm$ 0.5 & 15.6 $\pm$ 1.4
\\
& Ckpt Ens   & 32.6 $\pm$ 0.6 & 21.1 $\pm$ 0.4 & 8.4 $\pm$ 0.5 & 18.8 $\pm$ 0.9 & 9.1 $\pm$ 1.0

 \\
& Ensemble   & 27.1 $\pm$ 0.3 & 17.2 $\pm$ 1.3 & 10.0 $\pm$ 0.2 & 19.4 $\pm$ 0.6 & 10.1 $\pm$ 0.9

 \\
& Bayesian-LoRA(LA)      & 4.5 $\pm$ 0.3 & 9.6 $\pm$ 0.5 & 5.9 $\pm$ 0.2 & 8.5 $\pm$ 0.4 & 13.9 $\pm$ 1.2

 \\
& Bayesian-MoE(LA)      & \textbf{2.6} $\pm$ 0.2 & \textbf{8.4} $\pm$ 0.5 & \textbf{5.2} $\pm$ 0.1 & \textbf{7.0} $\pm$ 0.3 & \textbf{11.8} $\pm$ 1.5
 \\

\midrule

\multirow{6}{*}{NLL $\textcolor{red} \downarrow$}
& MAP        & 3.6 $\pm$ 0.1 & 2.7 $\pm$ 0.2 & 1.2 $\pm$ 0.1 & 1.7 $\pm$ 0.1 & 1.4 $\pm$ 0.1

 \\
& MC Drop    & 3.4 $\pm$ 0.1 & 2.3 $\pm$ 0.2 & 1.0 $\pm$ 0.1 & 1.6 $\pm$ 0.0 & 1.5 $\pm$ 0.1

 \\
& Ckpt Ens & 2.9 $\pm$ 0.2 & 2.0 $\pm$ 0.3 & 0.7 $\pm$ 0.1 & 1.0 $\pm$ 0.1 & 1.3 $\pm$ 0.0

 \\
& Ensemble  & 2.8 $\pm$ 0.2 & 1.9 $\pm$ 0.2 & 0.7 $\pm$ 0.1 & 1.3 $\pm$ 0.1 & 1.4 $\pm$ 0.0

 \\
& Bayesian-LoRA(LA) & 0.7 $\pm$ 0.0 & 0.7 $\pm$ 0.0 & \textbf{0.4} $\pm$ 0.1 & 0.7 $\pm$ 0.0 & \textbf{1.3} $\pm$ 0.1

 \\
& Bayesian-MoE(LA) & \textbf{0.6} $\pm$ 0.0 & \textbf{0.6} $\pm$ 0.0 & \textbf{0.4} $\pm$ 0.1 & \textbf{0.6} $\pm$ 0.0 & \textbf{1.1} $\pm$ 0.1

 \\

\bottomrule
\end{tabular}
\caption{
Out-of-distribution evaluation results for Bayesian-MoE applied to \textbf{DeepSeekMoE-16B-Base}. The model is fine-tuned on the OBQA dataset and evaluated on five distribution-shifted benchmarks.
}
\label{tab:bayesian-moe-results-4}
\end{table}
\subsection{Ablation: Layer-Wise Sensitivity of Bayesian-MoE}

To assess how different layers contribute to uncertainty estimation, we conduct an ablation study by selectively applying Bayesian-MoE to different portions of the model. Specifically, we divide the transformer layers of the LLM into four sequential quarters:

\begin{itemize}
    \item \textbf{Q1:} Layers $1$ to $L/4$
    \item \textbf{Q2:} Layers $L/4 + 1$ to $L/2$
    \item \textbf{Q3:} Layers $L/2 + 1$ to $3L/4$
    \item \textbf{Q4:} Layers $3L/4 + 1$ to $L$
\end{itemize}

We fine-tune the full model on the OBQA dataset and apply Bayesian-MoE post-hoc to the second linear layer of all experts \emph{except} for one excluded quarter. This setup allows us to examine how the absence of Bayesian treatment in different parts of the model impacts uncertainty calibration.

The results, reported in terms of Expected Calibration Error (ECE) and Negative Log-Likelihood (NLL), are shown for two model backbones: \texttt{Qwen1.5-MoE-A2.7B} and \texttt{DeepSeekMoE-16B-Base}.

\begin{table}[h]
\centering
\small
\begin{tabular}{lcc|cc}
\toprule
\textbf{Excluded Quarter} & \multicolumn{2}{c|}{\textbf{Qwen1.5-MoE-A2.7B}} & \multicolumn{2}{c}{\textbf{DeepSeekMoE-16B-Base}} \\
& ECE $\downarrow$ & NLL $\downarrow$ & ECE $\downarrow$ & NLL $\downarrow$ \\
\midrule
Exclude Q1 & 13.5 & 0.89 & 15.2 & 1.10 \\
Exclude Q2 &  9.6 & 0.75 & 10.1 & 0.83 \\
Exclude Q3 &  7.9 & 0.70 &  8.7 & 0.75 \\
Exclude Q4 &  6.6 & 0.65 &  7.1 & 0.69 \\
\bottomrule
\end{tabular}
\caption{Layer-wise ablation on Bayesian-MoE: Uncertainty metrics degrade more when earlier layers are excluded from Bayesian inference.}
\label{tab:bayesian-moe-layer-ablat}
\end{table}

The results in Table~\ref{tab:bayesian-moe-layer-ablat} indicate that earlier layers (especially Q1) contribute most significantly to calibrated uncertainty estimation. This suggests that the early expert modules encode critical uncertainty-relevant features, motivating their prioritization in memory- or compute-constrained Bayesian deployments.

\begin{figure}[t]
    \centering
    \includegraphics[width=0.95\textwidth]{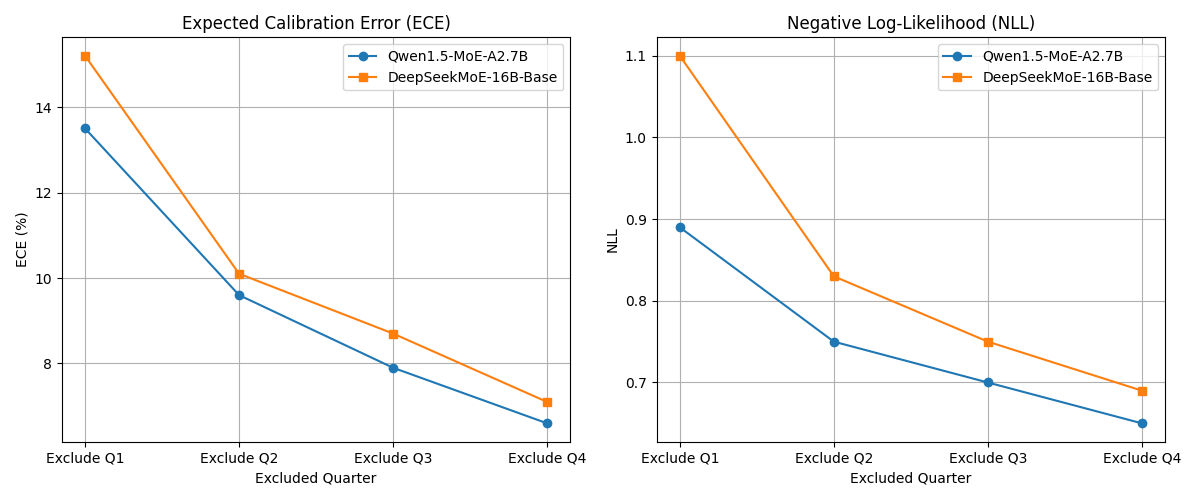}
    \caption{
    \textbf{Layer-wise Ablation of Bayesian-MoE on Qwen1.5-MoE-A2.7B and DeepSeekMoE-16B.}
    Each line shows the performance when excluding one of the four quarters of transformer layers from Bayesian-MoE treatment. The results indicate that earlier layers (Q1 and Q2) contribute more significantly to model calibration and likelihood. Notably, excluding the first quarter leads to the sharpest degradation in both Expected Calibration Error (ECE) and Negative Log-Likelihood (NLL), suggesting that early experts encode more uncertainty-relevant information. These trends are consistent across both model architectures.
    }
    \label{fig:layer-ablation}
\end{figure}

We also present the results in Figure \ref{fig:layer-ablation}, where each line shows the performance when excluding one of the four quarters of transformer layers from Bayesian-MoE treatment. The results indicate that earlier layers (Q1 and Q2) contribute more significantly to model calibration and likelihood. Notably, excluding the first quarter leads to the sharpest degradation in both Expected Calibration Error (ECE) and Negative Log-Likelihood (NLL), suggesting that early experts encode more uncertainty-relevant information. These trends are consistent across both model architectures.

\section{Conclusion}

We presented \textbf{Bayesian-MoE}, a modular and scalable Bayesian inference framework designed specifically for Mixture-of-Experts (MoE) large language models. By applying a structured Laplace approximation only to the second linear layers of the activated experts, our approach estimates posterior uncertainty in a memory-efficient manner using Kronecker-factored, low-rank approximations. Unlike Bayesian-LoRA, which introduces auxiliary adapter weights, Bayesian-MoE operates entirely over the existing MoE structure, preserving the model’s efficiency and sparsity. 

Empirical evaluations on Qwen1.5-MoE-A2.7B and DeepSeekMoE-16B demonstrate that Bayesian-MoE consistently improves uncertainty calibration (ECE) and predictive confidence (NLL) while remaining competitive in accuracy. This trend holds across both in-distribution and out-of-distribution benchmarks, underscoring the method’s robustness. Furthermore, Bayesian-MoE outperforms deep ensembles and other post-hoc methods while requiring only a single checkpoint. These results suggest that selectively Bayesianizing the expert layers of MoE LLMs is a principled and effective pathway for producing well-calibrated and reliable large language models. Future work will extend this framework to model expert correlations, apply it to other parts of the network such as routers or attention heads, and explore applications in open-ended generation tasks.
.

\clearpage
\bibliographystyle{unsrtnat}
\bibliography{references}

\clearpage
\appendix

\section{Posterior Estimation under Laplace Approximation}
Prior work \citep{yang2023bayesian} has evaluated several strategies for deriving predictive distributions under the Laplace approximation, including: (i) Monte Carlo sampling from the approximate Gaussian posterior, using either full or diagonal covariance; (ii) the probit approximation \citep{lu2020uncertainty,daxberger2021laplace}; and (iii) the Laplace bridge technique \citep{daxberger2021laplace,mackay1998choice}. Among these, Monte Carlo sampling with a full-covariance posterior was shown to yield the most accurate and calibrated predictions. Accordingly, we adopt this method for our Laplace-based predictive posterior computations, as detailed below.

\subsection{Monte Carlo Sampling from the Laplace Predictive Posterior}

Under the Laplace approximation, the posterior distribution over model parameters is Gaussian:
\[
\boldsymbol{\theta} \sim \mathcal{N}(\boldsymbol{\theta}_{\mathrm{MAP}},\, \boldsymbol{\Sigma})
\]
For a test input $\mathbf{x}_*$, this induces a Gaussian distribution over the model logits:
\begin{equation}
f_{\boldsymbol{\theta}}(\mathbf{x}_*) \sim \mathcal{N}\left(f_{\boldsymbol{\theta}_{\mathrm{MAP}}}(\mathbf{x}_*),\, \boldsymbol{\Lambda} \right)
\end{equation}
where the predictive covariance $\boldsymbol{\Lambda} \in \mathbb{R}^{|\mathcal{Y}| \times |\mathcal{Y}|}$ is given by
\begin{equation}
\boldsymbol{\Lambda} = \mathbf{J}(\mathbf{x}_*)^{\top} \boldsymbol{\Sigma} \mathbf{J}(\mathbf{x}_*),
\end{equation}
with Jacobian $\mathbf{J}(\mathbf{x}_*) := \nabla_{\boldsymbol{\theta}} f_{\boldsymbol{\theta}}(\mathbf{x}_*)\big|_{\boldsymbol{\theta} = \boldsymbol{\theta}_{\mathrm{MAP}}}$.

To sample from this Gaussian, we perform Cholesky decomposition (or stabilized factorization if needed) of the covariance:
\[
\boldsymbol{\Lambda} = \mathbf{L} \mathbf{L}^\top
\]
Letting $\boldsymbol{\xi} \sim \mathcal{N}(\mathbf{0}, \mathbf{I})$ be a standard normal vector, we define the sampling operator
\begin{equation}
\tilde{f}_{\boldsymbol{\theta}}(\mathbf{x}_*) = f_{\boldsymbol{\theta}_{\mathrm{MAP}}}(\mathbf{x}_*) + \mathbf{L} \boldsymbol{\xi}
\end{equation}

Monte Carlo estimates of the predictive distribution are then obtained by sampling $S$ i.i.d. replicates $\{\boldsymbol{\xi}^{(s)}\}_{s=1}^S$ and averaging the softmax outputs:
\begin{equation}
\mathbb{E}[\mathrm{softmax}(f_{\boldsymbol{\theta}}(\mathbf{x}_*))] \approx \frac{1}{S} \sum_{s=1}^{S} \mathrm{softmax}(\tilde{f}_{\boldsymbol{\theta}}^{(s)}(\mathbf{x}_*))
\end{equation}

This Monte Carlo sampling procedure is known to outperform deterministic approximations such as Probit or Laplace bridge, especially when using the full predictive covariance \citep{yang2023bayesian}. The method captures posterior uncertainty and yields well-calibrated predictive probabilities under the Laplace approximation.

\section{Algorithms}

\subsection{Efficient Low-Rank Factorization via Randomized SVD}

In large-scale Bayesian inference frameworks, the Fisher matrices and Jacobian-derived quantities can be extremely high-dimensional. When using structured Laplace approximations, we often need to estimate low-rank factorizations of empirical covariance matrices of the form:
\[
\mathbf{B}\mathbf{B}^\top \approx \sum_{t=1}^{T} \mathbf{b}_t \mathbf{b}_t^\top,
\]
where \(\mathbf{b}_t \in \mathbb{R}^d\) are either activation or gradient vectors collected over time.

However, performing a full singular value decomposition (SVD) over \(\mathbb{R}^{d \times T}\) becomes computationally expensive and memory intensive when both \(d\) and \(T\) are large (e.g., \(d > 10^4\)). To address this, we adopt a \emph{Randomized SVD (rSVD)} \citep{halko2011finding} approach that offers both scalability and approximation accuracy, making it well suited for iterative, large-batch scenarios encountered in model uncertainty quantification.

\begin{algorithm}[H]
\caption{Memory-efficient estimate of low-rank $\mathbf{B}$ using Randomized SVD}
\label{alg:lowrank-rsvd}
\begin{algorithmic}[1]
\State \textbf{Input:} number of steps $T$, target rank $n_{\text{kfac}}$, oversampling $p$
\State Initialize low-rank matrix $\mathbf{B} = \mathbf{0}$
\vspace{3pt}
\For{$t = 1, \dots, T$}
    \State Concatenate new observation $\mathbf{b}_t$ (e.g., $\mathbf{a}_t$ or $\mathbf{g}_t$) to $\mathbf{B} \leftarrow [\mathbf{B} \;\; \mathbf{b}_t]$
    \State Draw a random Gaussian matrix $\mathbf{\Omega} \sim \mathcal{N}(0, 1)^{d \times (n_{\text{kfac}} + p)}$
    \State Form sketch: $\mathbf{Y} = \mathbf{B} \mathbf{\Omega}$
    \State Orthonormalize: $\mathbf{Q} = \text{qr}(\mathbf{Y})$
    \State Project: $\mathbf{B}_{\text{small}} = \mathbf{Q}^\top \mathbf{B}$
    \State Compute SVD: $\mathbf{U}_{\text{small}}, \mathbf{S}, \mathbf{V}^\top = \text{svd}(\mathbf{B}_{\text{small}})$
    \State Update low-rank: $\mathbf{B} \leftarrow \mathbf{Q} \mathbf{U}_{\text{small}}[:, 1:n_{\text{kfac}}] \mathbf{S}[1:n_{\text{kfac}}; 1:n_{\text{kfac}}]$
\EndFor
\end{algorithmic}
\end{algorithm}

\subsection{Marginal Likelihood Optimization for Bayesian-MoE}
\label{marginal-liklihood-opt}
\begin{algorithm}[H]
\caption{Optimize Laplace Prior Precision for Bayesian-MoE Experts}
\label{alg:bayesian-moe-precision}
\begin{algorithmic}[1]
\State \textbf{Input:} Learning rate $\eta$, optimization steps $M$, initial prior precision $\lambda$
\State \textbf{Input:} Low-rank Kronecker factors $\{\mathbf{L}_a^{(e)}, \mathbf{L}_g^{(e)}\}_{e=1}^{E}$ for each expert
\State \textbf{Input:} MAP weights $\{\mathbf{W}_{\mathrm{MAP}}^{(e)}\}_{e=1}^{E}$
\vspace{3pt}
\For{$\text{step} = 1$ \textbf{to} $M$}
    \For{each expert $e = 1$ to $E$}
        \State Compute $\mathbf{C}_a^{(e)} = \mathbf{L}_a^{(e)\top} \mathbf{L}_a^{(e)}$
        \State Compute $\mathbf{C}_g^{(e)} = \mathbf{L}_g^{(e)\top} \mathbf{L}_g^{(e)}$
        \State Form $\mathbf{M}^{(e)} = \mathbf{I}_{r^2} + \lambda^{-1} \left( \mathbf{C}_a^{(e)} \otimes \mathbf{C}_g^{(e)} \right)$
        \State Compute $\log|\mathbf{H}^{(e)}| = d_{\text{in}} d_{\text{out}} \log \lambda + \log |\mathbf{M}^{(e)}|$
    \EndFor
    \State Compute total loss:
    \[
    \mathcal{L}_{\text{marg}}(\lambda) = \mathcal{L}(\mathbf{y}, \mathbf{X}; \boldsymbol{\theta}_{\mathrm{MAP}}) + \frac{1}{2} \sum_{e=1}^E \left( \log|\mathbf{H}^{(e)}| - \lambda \|\mathbf{W}_{\mathrm{MAP}}^{(e)}\|_F^2 \right)
    \]
    \State Update: $\lambda \gets \lambda + \eta \cdot \nabla_\lambda \mathcal{L}_{\text{marg}}(\lambda)$
\EndFor
\State \textbf{Return:} Optimized precision $\lambda$
\end{algorithmic}
\end{algorithm}

\subsection{Low-Rank Predictive Covariance Estimation for Bayesian-MoE}

Following the same steps as Bayesian-LoRA \citep{yang2023bayesian}, we use the Woodbury identity to compute the posterior precision matrix for each expert $e$:
\begin{align}
\boldsymbol{\Sigma}^{(e)}_{\text{post}} 
&= \left( (\mathbf{L}_a^{(e)} \otimes \mathbf{L}_g^{(e)})(\mathbf{L}_a^{(e)} \otimes \mathbf{L}_g^{(e)})^\top + \frac{1}{\sigma^2} \mathbf{I} \otimes \mathbf{I} \right)^{-1} \\
&= \sigma^2 \, \mathbf{I}_{d_{\text{in}}} \otimes \mathbf{I}_{d_{\text{out}}}
\; - \; \sigma^4 (\mathbf{L}_a^{(e)} \otimes \mathbf{L}_g^{(e)}) \, \mathbf{M}^{(e)\,-1} \, (\mathbf{L}_a^{(e)} \otimes \mathbf{L}_g^{(e)})^\top
\end{align}

where $\mathbf{M}^{(e)} = \mathbf{I}_{r^2} + \sigma^2 \left( \mathbf{C}_a^{(e)} \otimes \mathbf{C}_g^{(e)} \right)$ is the $r^2 \times r^2$ Woodbury correction matrix, and
\[
\mathbf{C}_a^{(e)} := \mathbf{L}_a^{(e)\top} \mathbf{L}_a^{(e)}, \quad \mathbf{C}_g^{(e)} := \mathbf{L}_g^{(e)\top} \mathbf{L}_g^{(e)}
\]

To compute the predictive variance for a new input $x_*$, we consider the covariance of the linearized function output:
\begin{equation}
\Lambda_{ij}^{(e)} = \mathbf{g}^{(e)\top}_i \, \boldsymbol{\Sigma}_{\text{post}}^{(e)} \, \mathbf{g}^{(e)}_j
\end{equation}

where $\mathbf{g}^{(e)}_i = \frac{\partial f_i(x_*)}{\partial \mathbf{W}^{(e)}} \in \mathbb{R}^{d_{\text{out}} \cdot d_{\text{in}}}$ is the gradient of the $i$-th output with respect to the parameters of the $e$-th expert's second linear layer.

Assuming a single layer, we reshape the gradient as:
\begin{equation}
\mathbf{g}^{(e)}_i = \mathrm{vec}(\mathbf{G}^{(e)}_i), \quad \text{where} \quad \mathbf{G}^{(e)}_i \in \mathbb{R}^{d_{\text{out}} \times d_{\text{in}}}
\end{equation}

Applying the Woodbury expression for $\boldsymbol{\Sigma}_{\text{post}}^{(e)}$, we get:
\begin{align}
\Lambda_{ij}^{(e)} &= \sigma^2 \, \mathrm{vec}(\mathbf{G}_i^{(e)})^\top \mathrm{vec}(\mathbf{G}_j^{(e)}) 
\; - \; \sigma^4 \, \mathrm{vec}(\mathbf{G}_i^{(e)})^\top (\mathbf{L}_a^{(e)} \otimes \mathbf{L}_g^{(e)}) 
\, \mathbf{M}^{(e)\,-1} \, (\mathbf{L}_a^{(e)\top} \otimes \mathbf{L}_g^{(e)\top}) \, \mathrm{vec}(\mathbf{G}_j^{(e)}) \\
&= \sigma^2 \, \langle \mathbf{G}_i^{(e)}, \mathbf{G}_j^{(e)} \rangle_F 
\; - \; \sigma^4 \, \mathrm{vec}(\mathbf{L}_g^{(e)\top} \mathbf{G}_i^{(e)} \mathbf{L}_a^{(e)})^\top 
\, \mathbf{M}^{(e)\,-1} \, \mathrm{vec}(\mathbf{L}_g^{(e)\top} \mathbf{G}_j^{(e)} \mathbf{L}_a^{(e)})
\end{align}

This final expression enables efficient computation of predictive uncertainty for each expert using only the low-rank approximations to the Kronecker factors, avoiding materializing large $d_{\text{out}} \times d_{\text{in}}$ matrices.

\subsection{Laplace Prior Optimization}
When validation data is available, we use a data-driven approach for maximizing the marginal likelihood: optimizing $\lambda$ using the validation log-likelihood. This method allows for more accurate control over predictive calibration by directly monitoring performance on unseen data. We follow the same approach and settings as described in Ref. ~\citep{yang2023bayesian}. Tables \ref{tab:laplace-prior-opt-results-1} and \ref{tab:laplace-prior-opt-results-2} provide the results of this experiment by splitting the training set into train/eval sets and applying Laplace prior optimization (LPO) using the eval set. As the results show, LPO could further improve the uncertainty of Bayesian-MoE in prediction.

\begin{table}[t]
\centering
\small
\setlength{\tabcolsep}{2pt}
\renewcommand{\arraystretch}{1.1}
\begin{tabular}{llcccccc}
\toprule
\textbf{Metric} & \textbf{Method} & \textbf{WG-S} & \textbf{ARC-C} & \textbf{ARC-E} & \textbf{WG-M} & \textbf{OBQA} & \textbf{MMLU} \\
\midrule

\multirow{2}{*}{ACC $\textcolor{green}\uparrow$}
& Bayesian-MoE(LA)           & 67.8 $\pm$ 0.1 & 66.4 $\pm$ 0.2 & 85.3 $\pm$ 0.5 & 73.9 $\pm$ 0.5 & 80.2 $\pm$ 0.2 & 62.5 $\pm$ 0.1 \\
& Bayesian-MoE(LA+LPO)           & 67.2 $\pm$ 0.1 & 65.7 $\pm$ 0.2 & 84.6 $\pm$ 0.5 & 73.1 $\pm$ 0.5 & 79.9 $\pm$ 0.2 & 62.2 $\pm$ 0.1 \\

\midrule

\multirow{2}{*}{ECE $\textcolor{red} \downarrow$}
& Bayesian-MoE(LA)       & 3.1 $\pm$ 0.2 & 7.5 $\pm$ 0.5 & 4.9 $\pm$ 0.1 & 6.5 $\pm$ 0.3 & 5.8 $\pm$ 0.5 & 11.5 $\pm$ 1.5 \\
& Bayesian-MoE(LA+LPO)       & \textbf{2.8} $\pm$ 0.2 & \textbf{6.7} $\pm$ 0.5 & \textbf{4.3} $\pm$ 0.1 & \textbf{6.1} $\pm$ 0.3 & \textbf{5.2} $\pm$ 0.4 & \textbf{9.6} $\pm$ 1.2 \\

\midrule

\multirow{2}{*}{NLL $\textcolor{red} \downarrow$}
& Bayesian-MoE(LA) & 0.6 $ \pm $ 0.01 &0.7 $\pm$ 0.01 &0.4 $\pm$ 0.06 &0.6 $ \pm $0.01 &0.6 $\pm$ 0.01 &1.3 $\pm$ 0.06 \\
& Bayesian-MoE(LA+LPO) & \textbf{0.4} $ \pm $ 0.02 &\textbf{0.5} $\pm$ 0.01 &\textbf{0.3} $\pm$ 0.04 &\textbf{0.4} $ \pm $0.02 &\textbf{0.3} $\pm$ 0.01 &\textbf{0.9} $\pm$ 0.04 \\

\bottomrule
\end{tabular}
\caption{
\textbf{Qwen1.5-MoE-A2.7B:} Zero-shot performance comparison of Bayesian-MoE coupled with Laplace prior optimization. Applying Laplace Prior Optimization (LPO) improves the performance.
}
\label{tab:laplace-prior-opt-results-1}
\end{table}


\begin{table}[t]
\centering
\small
\setlength{\tabcolsep}{2pt}
\renewcommand{\arraystretch}{1.1}
\begin{tabular}{llcccccc}
\toprule
\textbf{Metric} & \textbf{Method} & \textbf{WG-S} & \textbf{ARC-C} & \textbf{ARC-E} & \textbf{WG-M} & \textbf{OBQA} & \textbf{MMLU} \\
\midrule

\multirow{2}{*}{ACC $\textcolor{green}\uparrow$}
& Bayesian-MoE(LA)           & 70.8 $\pm$ 0.1 & 50.0 $\pm$ 0.2 & 68.6 $\pm$ 0.4 & 74.9 $\pm$ 0.5 & 81.7 $\pm$ 0.2 & 45.2 $\pm$ 0.1
\\
& Bayesian-MoE(LA+LPO)           & 69.1 $\pm$ 0.1 & 48.8 $\pm$ 0.3 & 67.3 $\pm$ 0.5 & 73.3 $\pm$ 0.4 & 79.2 $\pm$ 0.3 & 44.6 $\pm$ 0.2
\\

\midrule

\multirow{2}{*}{ECE $\textcolor{red} \downarrow$}
& Bayesian-MoE(LA)       & 2.5 $\pm$ 0.2 & 8.1 $\pm$ 0.5 & 4.9 $\pm$ 0.1 & 6.5 $\pm$ 0.3 & 5.9 $\pm$ 0.5 & 11.5 $\pm$ 1.5 \\
& Bayesian-MoE(LA+LPO)       & \textbf{1.8} $\pm$ 0.2 & \textbf{5.8} $\pm$ 0.4 & \textbf{3.2} $\pm$ 0.2 & \textbf{4.7} $\pm$ 0.4 & \textbf{4.2} $\pm$ 0.6 & \textbf{7.8} $\pm$ 1.2 \\

\midrule

\multirow{2}{*}{NLL $\textcolor{red} \downarrow$}
& Bayesian-MoE(LA) & 0.6 $\pm$ 0.0 & 0.6 $\pm$ 0.0 & 0.4 $\pm$ 0.1 & 0.6 $\pm$ 0.0 & 0.6 $\pm$ 0.0 & 1.0 $\pm$ 0.1
 \\
& Bayesian-MoE(LA+LPO) & \textbf{0.5} $\pm$ 0.1 & \textbf{0.4} $\pm$ 0.0 & \textbf{0.3} $\pm$ 0.1 & \textbf{0.4} $\pm$ 0.0 & \textbf{0.5} $\pm$ 0.1 & \textbf{0.7} $\pm$ 0.1
 \\

\bottomrule
\end{tabular}
\caption{
\textbf{DeepSeekMoE-16B-Base:} Zero-shot performance comparison of Bayesian-MoE coupled with Laplace prior optimization. Applying Laplace Prior Optimization (LPO) improves the performance.
}
\label{tab:laplace-prior-opt-results-2}
\end{table}

\section{Prompt Formats for Fine-Tuning}

To fine-tune both \texttt{Qwen1.5-MoE-A2.7B} and \texttt{DeepSeek-MoE-16B-Base}, we employ two standardized prompt templates corresponding to the two primary task formats: multiple-choice and binary (True/False) classification. These prompts are designed to align with instruction-following behavior and are kept consistent across all experiments.

\paragraph{Prompt Format for Multiple-Choice Questions.}
Each multiple-choice instance is structured as a natural language query followed by a list of labeled options. The model is asked to select the best answer from the provided choices:
\begin{quote}
\texttt{Choose the correct answer for the following question:} \\
\texttt{\{question\} Choices: A. \{option1\} B. \{option2\} C. \{option3\} D. \{option4\} Answer:}
\end{quote}

\paragraph{Prompt Format for True/False Questions.}
For binary classification, the input includes a question and an accompanying context passage. The model is expected to output a single word: \texttt{True} or \texttt{False}:
\begin{quote}
\texttt{Decide whether the statement is true or false.} \\
\texttt{Question: \{question\} Context: \{passage\}}
\end{quote}

\section{Fine-Tuning Setup for DeepSeek-MoE}

We fine-tune only the expert modules (MLPs) in the DeepSeek-MoE and Qwen1.5-MoE architecture, keeping the rest of the model weights frozen. This design reduces computational overhead and focuses learning capacity on the most adaptable parts of the model. Table~\ref{tab:moe-hparams} summarizes the hyperparameters used in our fine-tuning setup.

\begin{table}[ht]
\centering
\small
\renewcommand{\arraystretch}{1.2}
\begin{tabular}{ll}
\toprule
\textbf{Hyperparameter} & \textbf{Value / Description} \\
\midrule
Trainable Parameters       & Only MoE expert MLPs (e.g., second linear layer) \\
r (rank approximation)  &10 \\
Learning Rate              & $5 \times 10^{-5}$ \\
Learning Rate Scheduler    & Linear or Cosine \\
Weight Decay               & 0  \\
Dropout Probability        & 0.1 (applied within experts if used) \\
Batch Size                 & 32 \\
Max Sequence Length        & 256 \\
Number of Training Steps   & 10,000 \\
Optimizer                  & AdamW \\
Adam $\beta_1$, $\beta_2$  & (0.9, 0.95) or (0.9, 0.999) \\
Gradient Clipping          & 1.0 \\
Initial Loss Scale         & 65536 (for bf16 training) \\
\bottomrule
\end{tabular}
\caption{Fine-tuning hyperparameters for DeepSeek-MoE where only expert modules are updated.}
\label{tab:moe-hparams}
\end{table}

\section{Uncertainty Evaluation Metrics}

To assess the quality of uncertainty estimation, we employ two widely used metrics: \textit{Negative Log-Likelihood (NLL)} and \textit{Expected Calibration Error (ECE)}. These metrics quantify how well a model’s predicted confidence aligns with actual correctness.

\subsection{Negative Log-Likelihood (NLL)}

Negative Log-Likelihood measures the probabilistic correctness of model predictions. Given a model that outputs a categorical distribution over labels $p_{\boldsymbol{\theta}}(y \mid x)$, the NLL for a dataset $\{(x_i, y_i)\}_{i=1}^N$ is defined as:

\begin{equation}
\mathrm{NLL} = \frac{1}{N} \sum_{i=1}^{N} -\log p_{\boldsymbol{\theta}}(y_i==\hat{y}_i)
\end{equation}

where $y_i and \hat{y}_i$ are true and predicted outputs, respectively. Lower NLL values indicate that the model assigns higher probability to the correct labels. NLL penalizes both incorrect predictions and overconfident wrong predictions more severely, making it sensitive to miscalibrated uncertainty.

\subsection{Expected Calibration Error (ECE)}

ECE measures the discrepancy between predicted confidence and actual accuracy. It partitions predictions into $M$ confidence bins and computes the weighted average difference between confidence and accuracy in each bin. Formally:

\begin{equation}
\mathrm{ECE} = \sum_{m=1}^{M} \frac{|B_m|}{N} \left| \mathrm{acc}(B_m) - \mathrm{conf}(B_m) \right|
\end{equation}

where:
\begin{itemize}
    \item $B_m$ is the set of indices in the $m$-th confidence bin,
    \item $\mathrm{acc}(B_m)$ is the accuracy in bin $B_m$,
    \item $\mathrm{conf}(B_m)$ is the average predicted confidence in bin $B_m$:
\end{itemize}
\begin{equation}
\text{conf}(B_m) = \frac{1}{|B_m|} \sum_{i \in B_m} \mathbb{P}(\hat{y}_i),
\end{equation}

A lower ECE indicates better alignment between confidence and correctness, reflecting more calibrated predictions. Perfect calibration corresponds to $\mathrm{ECE} = 0$.


\section{Limitation}
While Bayesian-MoE  presents a promising approach to uncertainty estimation for Mixture-of-Experts (MoE) language models, it has several limitations. First, the method applies Bayesian inference solely to the second linear layer of each expert, and, therefore, the uncertainty arising from other parts of the model is not captured, potentially underestimating total epistemic uncertainty. Second, The posterior over expert parameters is assumed independent (block-diagonal), ignoring correlations between experts; therefore, in settings where experts are co-activated or their outputs are correlated, this simplification may harm calibration or likelihood estimates. Finally, the evaluation is conducted only on multiple-choice QA datasets, and the effectiveness of Bayesian-MoE on other tasks (e.g., open-ended generation, code synthesis, summarization) remains for future work.

\end{document}